\documentclass[letterpaper]{article} 
\usepackage{aaai25}  
\usepackage{times}  
\usepackage{helvet}  
\usepackage{courier}  
\usepackage[hyphens]{url}  
\usepackage{graphicx} 
\usepackage{natbib}  
\usepackage{caption} 
\usepackage{algorithm}
\usepackage{algorithmic}
\usepackage{url}  
\usepackage{booktabs}
\usepackage{amsfonts}  
\usepackage{nicefrac} 
\usepackage{microtype}   
\usepackage{fancyhdr}      
\usepackage{bm}
\usepackage{amsmath}
\usepackage{algorithmic}
\usepackage{algorithm}
\usepackage{multirow}
\usepackage{graphicx}
\usepackage{subfigure}
\usepackage{amsthm}
\usepackage{multicol}
\usepackage{pdfpages}
\usepackage{makecell}
\usepackage{xspace}
\usepackage{float}
\usepackage{threeparttable}
\usepackage{soul}
\usepackage{array}
\usepackage{dsfont}
\usepackage{pifont}
\usepackage{colortbl}
\usepackage{xcolor}

\usepackage{newfloat}
\usepackage{listings}
\DeclareCaptionStyle{ruled}{labelfont=normalfont,labelsep=colon,strut=off} 
\floatstyle{ruled}
\newfloat{listing}{tb}{lst}{}
\floatname{listing}{Listing}
\title{Multimodal Class-aware Semantic Enhancement Network for \\ Audio-Visual Video Parsing}
\newcommand*\samethanks[1][\value{footnote}]{\footnotemark[#1]}
\author {
    Pengcheng Zhao\equalcontrib,
    Jinxing Zhou\equalcontrib,
    Yang Zhao,
    Dan Guo\thanks{Corresponding authors.},
    Yanxiang Chen\samethanks
}
\affiliations {
    School of Computer Science and Information Engineering, Hefei University of Technology\\
    \{stevenzhao1001, zhoujxhfut\}@gmail.com, \{yzhao, guodan, chenyx\}@hfut.edu.cn
}

\usepackage{bibentry}

\begin{document}

\maketitle

\begin{abstract}
The Audio-Visual Video Parsing task aims to recognize and temporally localize all events occurring in either the audio or visual stream, or both.
Capturing accurate event semantics for each audio/visual segment is vital.
Prior works directly utilize the extracted holistic audio and visual features for intra- and cross-modal temporal interactions.
However, each segment may contain multiple events, resulting in semantically mixed holistic features that can lead to semantic interference during intra- or cross-modal interactions: the event semantics of one segment may incorporate semantics of unrelated events from other segments.
To address this issue, our method begins with a Class-Aware Feature Decoupling (CAFD) module, which explicitly decouples the semantically mixed features into distinct class-wise features, including multiple event-specific features and a dedicated background feature.
The decoupled class-wise features enable our model to selectively aggregate useful semantics for each segment from clearly matched classes contained in other segments, preventing semantic interference from irrelevant classes.
Specifically, we further design a Fine-Grained Semantic Enhancement module for encoding intra- and cross-modal relations. 
It comprises a Segment-wise Event Co-occurrence Modeling (SECM) block and a Local-Global Semantic Fusion (LGSF) block.
The SECM exploits inter-class dependencies of concurrent events within the same timestamp with the aid of a new event co-occurrence loss. The LGSF further enhances the event semantics of each segment by incorporating relevant semantics from more informative global video features.
Extensive experiments validate the effectiveness of the proposed modules and loss functions, resulting in a new state-of-the-art parsing performance.
\end{abstract}

%

\section{Introduction}\label{sec:introduction}
\begin{figure}[t]
  \centering
\includegraphics[width=0.95\linewidth]{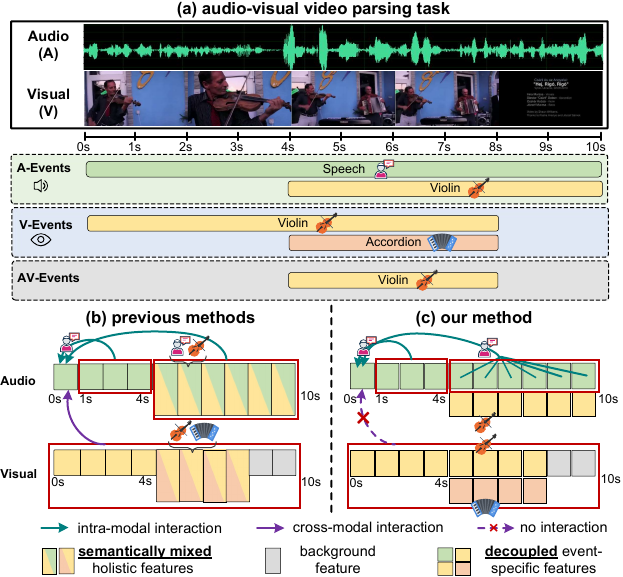}
  \caption{(a) Illustration of the AVVP task. (b) Previous methods rely on semantically mixed holistic audio/visual features for intra- and cross-modal interactions, leading to semantic interference.
  (c) In contrast, we utilize decoupled class-aware features to aggregate event semantics for each segment from only matched classes during interactions. 
  }
  \label{fig:introduction}
\end{figure}

In this paper, we focus on the task of Audio-Visual Video Parsing (AVVP) for the fundamental understanding of audio-visual scenes.
As illustrated in Fig.~\ref{fig:introduction} (a), given an audible video, the AVVP task seeks to temporally localize all events of interest, including audio events, visual events, and audio-visual events (both audible and visible).
Notably, the audio and visual signals are not required to be temporally aligned in this task.
As shown in Fig.~\ref{fig:introduction} (a), the event \textit{Speech} is present only in the audio modality while the \textit{Accordion} is solely present in the visual modality.
Moreover, multiple events can overlap on the timeline.
To discern various types of events, an AVVP model needs to \textit{accurately perceive event semantics for each audio/visual segment}.
Given the multimodal and temporal characteristics of this task, the model must be carefully designed to not only utilize event semantics shared across audio and visual modalities but also maintain the unique event semantics inherent to each modality. 

Previous works~\cite{tian2020unified,yu2021mm,lai2023modality} have progressed this task by developing various sophisticated network architectures.
For instance, one representative work, HAN~\cite{tian2020unified}, introduces a hybrid attention network that utilizes self-attention and cross-attention mechanisms to capture event semantics through the modeling of temporal relations within intra- and cross-modalities, enabling each temporal segment to interact with all segments from the same or the other modality.
Although these methods achieve impressive video parsing performances, they directly operate on \textit{semantically mixed} audio and visual holistic features extracted by pretrained audio/image classification  backbones~\cite{hershey2017vggish,he2016resnet}, which could lead to \textbf{semantic interference} during intra- and cross-modal temporal modeling.
Specifically, as illustrated in Fig.~\ref{fig:introduction} (b), the first audio segment only contains the event {`Speech'}.
During intra-modal temporal modeling, interactions between this segment and those audio segments at (1s, 4s] will undoubtedly be beneficial for enhancing its event semantic of `Speech' because they share the identical single `Speech' event; however, this segment will incorporate the unrelated semantic of event `Violin' when interacting with those audio segments at (4s, 10s], causing intra-modal semantic interference.
Similarly, semantic interference can also occur during the cross-modal temporal modeling when the audio segment interacts with visual segments containing only partially similar or completely different events, \textit{e.g.}, the interactions between the first audio segment and all visual segments in the example shown in Fig.~\ref{fig:introduction} (b).
To address this issue, we propose investigating \textit{whether we can enhance the event semantics of each audio/visual segment by selectively aggregating only the most closely matched semantics from other segments during the intra-modal and cross-modal interactions}.

To this end, we first propose a \textbf{Class-Aware Feature Decoupling (CAFD)} module to explicitly decouple the semantically mixed holistic features of each audio/visual segment into distinct class-wise features, allowing our model to perform the interactions at a more fine-grained and precise class level.
Specifically, the feature decoupling can be implemented by employing multiple independent linear layers on the holistic audio/visual feature for each segment.
Particularly, in addition to the \textit{event-specific} classes, we include a special \textit{background} class in our feature decoupling process.
This design is motivated by two observations: 1) each segment contains a degree of background information, and some background context may facilitate the event recognition. For example, the background of {parking lot} suggests that the event may relate to the cars; 
2) some segments may contain only useless background noise, \textit{e.g.}, the last two visual segments in Fig.~\ref{fig:introduction} (a). 
Features of such segments should not be decoupled into other event-specific classes. 
Considering these observations, we employ a weighting mechanism to dynamically blend the background feature into the event-specific features to better utilize the decoupled background information. 
Notably, we introduce a reconstruction loss and an orthogonality loss to guarantee effective feature decoupling. 

Then, we propose a \textbf{Fine-Grained Semantic Enhancement (FGSE)} module to fully exploit the decoupled class-aware audio and visual features.  
The FGSE module comprises a \textit{Segment-wise Event Co-occurrence Modeling (SECM)} block and a \textit{Local-Global Semantic Fusion (LGSF)} block.
The SECM is designed to model inter-class dependencies among concurrent events within the same timestamp, enabling each event-specific feature to aggregate useful semantics from other relevant classes, 
where we also propose a novel \textit{event co-occurrence loss function} to facilitate the learning of the event co-occurrence map. 
And the LGSF block considers event relations across temporal stamps by interacting the feature of each local segment with the global video feature, which is derived by averaging all the temporal features to provide more informative and noise-robust event semantics as events typically span consecutive temporal segments. 
Notably, the SECM and LGSF blocks can be applied to both intra-modality and cross-modality. More implementation details will be introduced in Sec.~\ref{sec:FGSE}.

Our main contributions can be summarized as follows:
\begin{itemize}
\item We analyze the intractable semantic interference issue in the AVVP task and present a novel Multi-Modal Class-aware Semantic Enhancement (MM-CSE) network.
\item We propose a 
class-aware feature decoupling 
module to decouple the semantically mixed audio/visual features into distinct event-specific and background features, facilitating event semantic perception from a more precise class level. 
\item We propose a 
fine-grained semantic enhancement 
module consisting of a segment-wise event co-occurrence block and a local-global semantic fusion block, further enhancing the event semantics of each segment. A new event co-occurrence loss function is introduced in this module.
\item Extensive quantitative and qualitative experiments validate the effectiveness of our method, which achieves a new state-of-the-art audio-visual video parsing performance.
\end{itemize}
\begin{figure*}[t]
  \centering
  \includegraphics[width=0.96\textwidth]{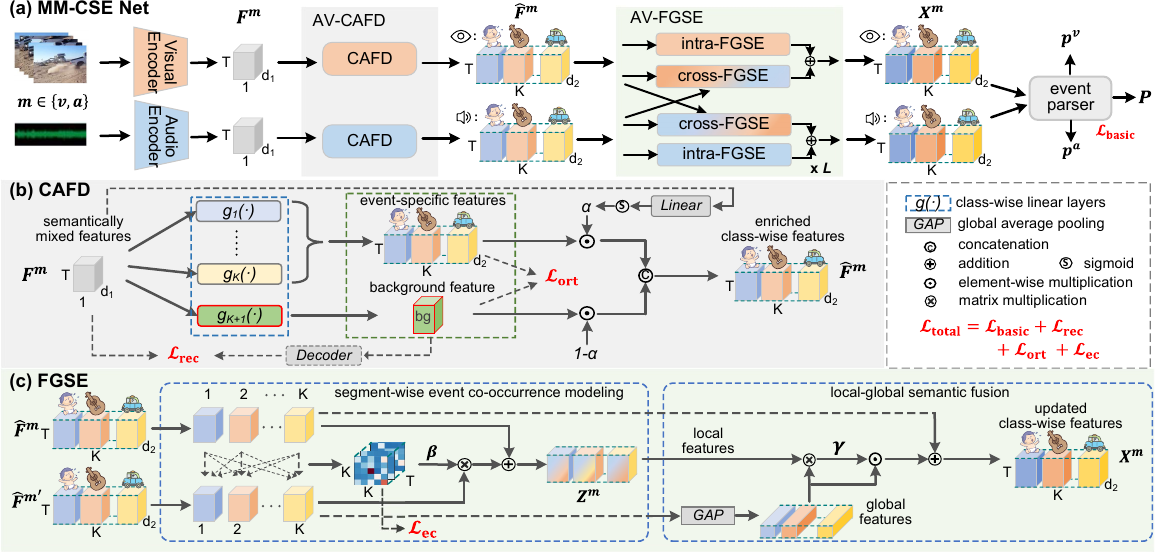}
  \caption{Framework Overview. 
  (a) Our MM-CSE network primarily consists of two core modules: the audio-visual Class-Aware Feature Decoupling (AV-CAFD) and the Fine-Grained Semantic Enhancement (AV-FGSE).
 (b) The CAFD module decouples the encoded audio/visual features into distinct class-wise features, each representing a specific event or the background class.
 To ensure effective decoupling, we introduce a reconstruction loss $\mathcal{L}_\text{rec}$ and an orthogonality loss $\mathcal{L}_\text{ort}$.
 (c) The FGSE module further enhances the obtained class-wise features through two successive blocks: the Segment-wise Event Co-occurrence Modeling (SECM) block and the Local-Global Semantic Fusion (LGSF) block.
 The SECM encodes the relations among concurrent events within each timestamp, whereas the LGSF enhances the event semantics of local temporal segments by fusing relevant semantics of the global video.
 We also introduce an event co-occurrence loss $\mathcal{L}_\text{ec}$ to steer the learning of event co-occurrence in the SECM block.
 Notably, the FGSE module is applied to both intra-modality (`intra-FGSE' ) and cross-modality (`cross-FGSE').
 }
  \label{fig:framework}
\end{figure*}

\section{Related Work}
\textbf{Audio-Visual Learning}
aims to leverage both audio and visual modalities to enable machines to emulate the human perception process~\cite{cheng2020look,mao2024tavgbench,shen2023fine,li2024object,zhou2024label}. 
Tasks such as sound source localization~\cite{zhao2018sound,qian2020multiple}, 
audio-visual segmentation~\cite{zhou2022avs,zhou2023avss,guo2023audio}
and event localization~\cite{tian2018audio,zhou2021positive,zhou2022cpsp,zhou2024towards} explore audio-visual learning from spatial or temporal perspectives.
Most of these prior studies presume synchronized audio and visual signals to be semantic-corresponding. 
However, this assumption does not hold in many situations, such as when the sound source is off-screen.
In our studied AVVP task, the audio and visual signals may not always be temporally aligned. Our goal is to discern events within each modality, necessitating a more robust approach for articulating both intra-modal and inter-modal relations.

\noindent\textbf{Audio-Visual Video Parsing}
aims to comprehensively identify and temporally localize events occurring in independent audio and visual modalities.
In early research works~\cite{tian2020unified}, this task is approached under a weakly supervised setting, where only the event label of the entire video is available for model training.
To enhance supervision, some subsequent works have attempted to generate pseudo labels of events for independent audio and visual modalities at video-level~\cite{wu2021exploring,cheng2022joint} or more fine-grained segment-level~\cite{zhou2023improving,lai2023modality,fan2023revisit,zhou2024vaplan}. 
The generated segment-level pseudo labels can significantly improve the video parsing performance.
Moreover, a variety of methods are proposed to encode more effective audio-visual representations for superior event parsing. 
The pioneer work HAN~\cite{tian2020unified} proposes a hybrid attention network that adopts the multi-head attention~\cite{vaswani2017attention} mechanism to encode the intra-modal and cross-modal relations.
Subsequently, 
some methods~\cite{yu2021mm,jiang2022dhhn,zhang2023multi} are specially designed to better localize audio/visual events at varied temporal durations.
To identify multiple events in audio or visual modalities, MGN~\cite{mo2022multi} utilizes multiple learnable class tokens to group the event semantics contained in each modality.
Our method is significantly different from MGN as well as previous methods because we aim to address the semantic interference and our model relies on the decoupled class-wise features rather than the semantically mixed hidden features.
More importantly, our approach uniquely accounts for a special \textit{background} class, which is overlooked in all prior methods. 

\section{Our Approach}
Given a video sequence comprising $T$ non-overlapping pairs of audio and visual segments  $\left\{a_t,\ v_t\right\}_{t=1}^T$, the AVVP task aims to predict all types of events within each segment pair, labeled as $\left(y_t^a, y_t^v, y_t^{av}\right) \in \mathbb{R}^{1\times K}$.
Here, $y_t^a$, $y_t^v$ and $y_t^{av}=y_t^a\ast y_t^v$ denote the audio, visual and audio-visual events, respectively, and 
$K$ is the total number of event categories.
Notably, each segment may contain multiple events or no events at all.
During training, we can only access the manually annotated video-level label $\bm{Y} \in\mathbb{R}^{1\times K}$, depicting all events present in the video.
However, the weak label $\bm{Y}$ does not specify the temporal segments or the modality in which these events occur.
The most recent work~\cite{lai2023modality} provides high-quality pseudo labels for each modality at the segment-level, denoted as $\bm{\hat{y}}^a, \bm{\hat{y}}^v \in \mathbb{R}^{T \times K}$, offering fine-grained supervision. 
Fig.~\ref{fig:framework} illustrates the framework of our proposed MM-CSE. We provide its details in the next subsections.

\subsection{Class-Aware Feature Decoupling}\label{sec:CAFD}

In the AVVP task, each audio/visual segment may contain multiple events; consequently, the event semantics are intermixed in the extracted audio/visual feature $\bm{F}^a/\bm{F}^v \in \mathbb{R}^{T \times d_1}$, resulting in semantic interference when performing intra-modal and cross-modal interactions, as discussed in Sec.~\ref{sec:introduction}.

To address this issue, we propose a CAFD 
module to explicitly decouple the semantically mixed features $\bm{F}^a, \bm{F}^v \in \mathbb{R}^{T\times d_1}$ into distinct class-aware features $\widetilde{\bm{F}}^a$, $\widetilde{\bm{F}}^v \in \mathbb{R}^{T \times (K+1) \times d_2}$, with `$K+1$' denoting the $K$ \textit{event} class plus an additional \textit{background} class.
We incorporate a background class for each segment to account for both the event-related background context and segments that may only contain non-informative noise. 

As shown in Fig.~\ref{fig:framework} (b), we accomplish the feature decoupling via $K$+1 independent linear layers $\{g_k\}_{k=1}^{K+1}$,  
obtaining the decoupled features $\widetilde{\bm{F}}^m =\{ \widetilde{\bm{f}}_k^m \}_{k=1}^{K+1}$, where $m \in \{a, v\}$ denotes the audio and visual modalities, respectively. 
Here, we define the features $\{ \widetilde{\bm{f}}_k^m \}_{k=1}^{K}$
as the event-specific features $\widetilde{\bm{F}}_\text{e}^m$ and $\widetilde{\bm{f}}_{K+1}^m$
as the background feature $\widetilde{\bm{F}}_\text{bg}^m$. 
It is worth noting that we introduce a reconstruction loss $\mathcal{L}_\text{rec}$ and an orthogonality loss $\mathcal{L}_\text{ort}$ on the decoupled features to guide effective feature decoupling. 
The former helps to maximize the semantic preservation of decoupled class-wise features compared to the original holistic feature, while the latter minimizes the relevance between the event-specific and the background features to make them more discriminative.
We will elaborate on these loss functions in Sec.~\ref{sec:loss_functions}.

Considering some background information could also be helpful for event recognition, e.g., the background context may indicate this is an indoor event, we further dynamically integrate the background feature $\widetilde{\bm{F}}_\text{bg}^m$ into each event class feature $\widetilde{\bm{F}}_\text{e}^m$.
Specifically, we first utilize linear layers and the Sigmoid function to generate a fusion weight vector $\alpha^m \in \mathbb{R}^{T\times 1}$ through the original holistic feature $\bm{F}^m$. 
In principle, $\alpha^m$ indicates the importance of the background in recognizing the current segments.
Then, we employ a weighting mechanism to blend $\widetilde{\bm{F}}_\text{bg}^m$ with $\widetilde{\bm{F}}_\text{e}^m$, yielding the enriched class-wise feature $\widehat{\bm{F}}^m$ as follows,
\begin{equation}
    \widehat{\bm{F}}^m = h[\alpha^m \cdot \widetilde{\bm{F}}_\text{e}^m; (1-\alpha^m) \cdot \widetilde{\bm{F}}_\text{bg}^m],
\label{eq:CAFD_add_bg_event}
\end{equation}

\noindent where $\widehat{\bm{F}}^m \in \mathbb{R}^{T \times K \times d_2}$, $h$ is implemented by a linear layer followed by the ReLU function. 
Note that $\widetilde{\bm{F}}_\text{bg}^m$ is repeated $K$ times to match the dimension of $\widetilde{\bm{F}}_\text{e}^m $ for the concatenation operation $[;]$ in Eq.~\ref{eq:CAFD_add_bg_event}, and we omit this for simplicity.

\subsection{Fine-Grained Semantic Enhancement}\label{sec:FGSE}
After obtaining the enriched class-wise features $\widehat{\bm{F}}^m \in \mathbb{R}^{T \times K \times d_2}, m\in\{a, v\}$,
we consider enhancing event semantics of each segment by exploring the semantic relevance from a more fine-grained class-level.
To this end, we propose a FGSE
module.
Specifically, given the class-wise audio feature $\widehat{\bm{F}}^a$ and visual feature $\widehat{\bm{F}}^v$, our FGSE module encodes the intra-modal and cross-modal interactions as follows:
\begin{equation}
\bm{X}^m = \varphi_\text{intra}^m(\widehat{\bm{F}}^m, \widehat{\bm{F}}^m) + \varphi_\text{cross}^m(\widehat{\bm{F}}^m, \widehat{\bm{F}}^{\bar{m}}),
\label{eq:FGSE_self_cross}
\end{equation}

\noindent where $\bm{X}^m \in \mathbb{R}^{T \times K \times d_2}$ is the updated feature for the $m$ modality, and $\bar{m}=\{a, v\} \backslash m$ denotes the counterpart modality.
As illustrated in Fig.~\ref{fig:framework} (a) and (c), $\varphi_\text{intra}$ and $\varphi_\text{cross}$ share the same operations but have different parameters, each consisting of a 
\textit{SECM} block and a \textit{LGSF} block.
We introduce the design principles and details of these two blocks next.

\textbf{Segment-wise Event Co-occurrence Modeling (SECM).}
Each audio/visual segment may contain multiple events, indicating the event co-occurrence.
For example, events such as 
`people cheering' and `people clapping' 
frequently co-occur within or across modalities, highlighting their relevant event semantics. Given that the class-wise features $\widehat{\bm{F}}^m 
\in \mathbb{R}^{T \times K \times d_2}$ 
are available, we design the SECM block to exploit the dependencies of such concurrent events.
Specifically, for each segment, we measure the event co-occurrence by accessing the similarity~\cite{vaswani2017attention} among $K$-class event features, computed as follows,
\begin{equation}
    \bm{\beta}^{m,m'} = \text{Softmax}(\widehat{\bm{F}}^m (\widehat{\bm{F}}^{m^{'}})^{\top}/{\sqrt{d_2}}),
\end{equation}

\noindent where $m'\in\{m,\bar{m}\}$. 
The resulting $\bm{\beta}^{m,m'} \in \mathbb{R}^{T \times K \times K}$ indicates the event co-occurrence weight, where a larger value $\bm{\beta}^{m,m'}_{:,i,j}$ implies that the $i$-th event in modality $m$  is more likely to coexist with the $j$-th event in modality $m'$.
Then, each event-specific feature can be enhanced by aggregating relevant semantics from features of its concurrent events, computed by, $\bm{Z}^m = \widehat{\bm{F}}^m + \bm{\beta}^{m,m'} (\widehat{\bm{F}}^{m'})^{\top}$,
where $\bm{Z}^m \in \mathbb{R}^{T \times K \times d_2}$ is the enhanced class-wise feature. 

\textbf{Local-Global Semantic Fusion (LGSF).}
The above SECM block performs inter-class co-occurrence modeling \textit{within each local timestamp}.
In order to harness relevant event semantics \textit{across temporal timestamps}, we consider enhancing the semantics of each class within local segments by utilizing the global video information.
We consider such a local-global interaction recognizing that the event semantics of the global video are typically more robust and informative, given that an event often spans consecutive temporal timestamps.

Specifically, we obtain the event feature of global video $\overline{\bm{G}}^{m'} \in \mathbb{R}^{1 \times K \times d_2}$ via Global Average Pooling of $\widehat{\bm{F}}^{m'}$ along the temporal dimension. 
Then, we evaluate the semantic relevance between each local segment and the global video by computing their cosine similarity $\bm{\gamma}^{m,m'}$: 
\begin{equation}
    \bm{\gamma}^{m,m'} = \langle \| \bm{Z}^{m} \|, \|\overline{\bm{G}}^{m'} \| \rangle,
\end{equation}

\noindent where $\| \cdot \|$ denotes the L2-normalization, $\bm{\gamma}^{m,m'} \in \mathbb{R}^{T \times K \times 1}$. The element $\bm{\gamma}^{m,m'}_{:,k} \in \mathbb{R}^{T \times 1}$ indicates the feature similarity between $T$ local segments with the global video for the $k$-th event class.
Afterward, the feature of each local segment can be enhanced by aggregating relevant event semantics from global video feature via $\bm{\gamma}^{m,m'}$, calculated as, $\bm{X}^m = \widehat{\bm{F}}^m + \bm{\gamma}^{m,m'} \odot \overline{\bm{G}}^{m'}$. 

For convenience, we abbreviate the above operations of the SECM and the LGSF blocks as a single FGSE layer: 
\begin{equation}
    \bm{X}^a, \bm{X}^v = \text{FGSE}(\widehat{\bm{F}}^a, \widehat{\bm{F}}^v),
\end{equation}

\noindent Then, we utilize 
$L$ stacked FGSE layers to iteratively enhance the class-wise audio and visual features:
\begin{equation}
    \bm{X}^a_{l}, \bm{X}^v_{l} = \text{FGSE}({\bm{X}}^a_{l-1}, {\bm{X}}^v_{l-1}),
\end{equation}

\noindent where $l=\{1,2,...,L\}$ is the layer index and $\bm{X}^m_0=\widehat{\bm{F}}^m$.

\subsection{Loss Functions}\label{sec:loss_functions}
The audio and visual features obtained by the final FGSE layer, $\bm{X}^m_L \in \mathbb{R}^{T \times K \times d_2} (m\in\{a,v\})$, are fed into the event parser to predict the segment-level event probabilities $\bm{p}^m \in \mathbb{R}^{T \times K}$.
The video-level prediction for the entire video $\bm{P} \in \mathbb{R}^{1 \times K}$ can be obtained from $\bm{p}^a$ and $\bm{p}^v$ through an attentive multi-modal multi-instance (MMIL) pooling~\cite{tian2020unified} mechanism.

\textbf{Basic classification loss $\mathcal{L}_\text{basic}$.} 
Given the segment-level and video-level predictions, the basic loss function $\mathcal{L}_\text{basic}$ utilizes the binary cross entropy (BCE) loss to align them with the ground truths, computed as,
\begin{footnotesize}
\begin{equation}
\mathcal{L}_\text{basic} = \text{BCE}(\bm{p}^a, \bm{\hat{y}^a}) + \text{BCE}(\bm{p}^v, \bm{\hat{y}^v}) + \text{BCE}(\bm{P}, \bm{Y}),
\end{equation}
\end{footnotesize}

Additionally, our optimization objective involves two loss functions aimed at facilitating effective class-aware feature decoupling (\textbf{$\mathcal{L}_\text{rec}$} and \textbf{$\mathcal{L}_\text{ort}$}) and a novel loss function \textbf{$\mathcal{L}_\text{ec}$} designed to regularize the event co-occurrence modeling in the SECM module. Next, we elaborate on each of them.

\textbf{Reconstruction loss $\mathcal{L}_\text{rec}$}. 
To maximize the preservation of semantic information within the decoupled features, we feed the decoupled class-wise features $\widetilde{\bm{F}}^m$ into a \textit{Decoder} to reconstruct the original holistic feature ${\bm{F}}^m$. The \textit{Decoder} is implemented by two linear layers inserted by a ReLU activation. 
Then, we calculate the reconstruction loss $\mathcal{L}_\text{rec}$ by the mean squared error (MSE):
\begin{equation}
\mathcal{L}_\text{rec}= \sum_{m} \text{MSE}(\textit{Decoder}(\widetilde{\bm{F}}^m), {\bm{F}}^m ),
\label{eq:loss_rec}
\end{equation}

\textbf{Orthogonality loss $\mathcal{L}_\text{ort}$.} In addition to $\mathcal{L}_\text{rec}$, $\mathcal{L}_\text{ort}$ is introduced to promote dissimilarity between the background feature $\widetilde{\bm{F}}_\text{bg}^m \in \mathbb{R}^{T \times 1 \times d_2}$ and the event-specific feature $\widetilde{\bm{F}}_\text{e}^m \in \mathbb{R}^{T \times K \times d_2}$, enhancing the isolation of the background information for improved event prediction.
$\mathcal{L}_\text{ort}$ is calculated as follows,
\begin{equation}
\mathcal{L}_\text{ort}=\frac{1}{TK}\sum_{m}\sum_{t=1}^{T}\sum_{k=1}^{K}{ \langle \| \widetilde{\bm{F}}_\text{bg}^m \|, \| \widetilde{\bm{F}}_\text{e}^m } \| \rangle,
\label{eq:loss_ort}
\end{equation}

\noindent where $\langle \| \widetilde{\bm{F}}_\text{bg}^m \|, \| \widetilde{\bm{F}}_\text{e}^m \| \rangle$ is the cosine similarity matrix which has the dimension of $T \times K$.
 
\textbf{Event co-occurrence loss $\mathcal{L}_\text{ec}$.}
Within the SECM module, the $\bm{\beta}^{m,m'} \in \mathbb{R}^{T \times K \times K}$ reflects the learned event co-occurrence relations between modality $m$ and modality $m'$.
We can also obtain the event co-occurrence matrix $\bm{M}^{m,m'} \in \mathbb{R}^{T \times K \times K}$ from the segment-level labels $\bm{\hat{y}}^{m}, \bm{\hat{y}}^{m'} \in \mathbb{R}^{T \times K}$ as ground truth, i.e.,
\begin{equation}
\bm{M}^{m,m^\prime}_{t,(i,j)}=\begin{cases}
1, &\text{ if } {\bm{\hat{y}}^m_{t,i}}={\bm{\hat{y}}^{m'}_{t,j}}=1 \\
0, & other
\end{cases}
\end{equation}

\noindent Then, the $\mathcal{L}_\text{ec}$ can be computed by,
\begin{equation}
    \mathcal{L}_\text{ec} = \text{MSE}(\bm{\beta}^{m,m'}, \bm{M}^{m,m'}),
\end{equation}

\textbf{In summary}, the overall objective $\mathcal{L}_\text{total}$ is defined as:
\begin{equation}
\mathcal{L}_\text{total} =\mathcal{L}_\text{basic}+\mathcal{L}_\text{rec}+\lambda_1\mathcal{L}_\text{ort}+\lambda_2\mathcal{L}_\text{ec}.
\label{eq:loss_total}
\end{equation}
where $\lambda_1$ and $\lambda_2$ are the balancing hyperparameters.

\section{Experiments}

\subsection{Experimental Setups}
\textbf{Dataset \& Metrics.} Following prior works~\cite{jiang2022dhhn,lai2023modality}, 
our experiments are conducted on the \textit{Look, Listen, and Parse} (LLP)~\cite{tian2020unified} dataset, which is currently the sole standard dataset used for the AVVP task.
It comprises 11,849 videos covering 25 common categories. Following the official data splits, 
the dataset is divided into 10,000 videos for training, 649 for validation, and 1,200 for testing. 
For the evaluation metrics, we adopt F-scores across all types of events: audio events ({A}), visual events ({V}), and audio-visual events ({AV)}, at both segment-level and event-level. 
Furthermore, `Type' is the averaged F-scores of A,
V, and AV metrics. `Event' considers audio and visual
events simultaneously for each sample. 

\noindent\textbf{Implementation details.} \textit{1) Feature extraction.} 
Following the previous SOTA method VALOR, video frames are sampled at 8 FPS and the pretrained CLIP~\cite{radford2021CLIP} and R(2+1)D~\cite{tran2018closer} models are utilized to extract the 2D and 3D features, respectively, which are then concatenated to form the visual features.
The corresponding audio features are extracted using the pretrained CLAP~\cite{wu2023clap} model. 
\textit{2) Training configurations.} Our model is trained for 60 epochs with a batch size of 64 using AdamW 
optimizer, with an initial learning rate of 3e-4 and a weight decay of 1e-3. 
Feature dimensions $d_1$ and $d_2$ are set to 256 and 128, respectively. We use $L=4$ stacked FGSE layers. The hyperparameters $\lambda_1$ and $\lambda_2$ in Eq.~\ref{eq:loss_total} are empirically set to 0.1. 
The code will be publicly available.
\begin{table*}[!t]
\small
  \centering
  \label{tab1}
\setlength{\tabcolsep}{1mm}
\begin{tabular}{c|ccccc|ccccc|c}
\Xhline{1.5pt}
\multirow{2}{*}{Methods} & \multicolumn{5}{c|}{Segment-level}
 & \multicolumn{5}{c|}{Event-level} & \multirow{2}{*}{Avg.}                                       \\ 
& A & V & AV & Type  & Event                                     & A  & V    & AV  & Type  & Event    &                                                            \\ \hline
HAN~\cite{tian2020unified}             & 60.1  & 52.9  & 48.9   & {54.0}   & 55.4    & 51.3  & 48.9     & 43.0                                                       & 47.7     & 48.0  & 51.0                                                       \\ 
CVCMS~\cite{lin2021exploring}         & 59.2                                                & 59.9                                                      & 53.4                                                   & 57.5                                                    & 58.1                                                       & 51.3             & 55.5                                                    & 46.2                                                     & 51.0                                                   & 49.7                                                       & 54.2                                                       \\
MA~\cite{wu2021exploring}            & {60.3}                                                       & {60.0}                                                       & {55.1}                                                       & {58.9}                                                       & 57.9                                                       & {53.6}                                                       & {56.4}                                                       & {49.0}                                                       & {53.0}                                                       & 50.6                                                       & 55.5                                                       \\
MGN~\cite{mo2022multi}            & 60.2 & 61.9 & 55.5 & 59.2 & 58.7 & 50.9 & 59.7 & 49.6 & 53.4 & 49.9 & 55.9                     \\ 
MM-Pyr~\cite{yu2021mm}           & 61.1 & 60.3 & 55.8 & 59.7 & 59.1 & 53.8 & 56.7 & 49.4 & 54.1 & 51.2 & 56.1                                                       \\ 
JoMoLD~\cite{cheng2022joint}          & {61.3}                                                       & {63.8}                                                       & {57.2}                                                       & {60.8}                                                       & 59.9                                                       & {53.9}                                                       & {59.9}                                                       & {49.6}                                                       & {54.5}                                                       & 52.5                                                       & 57.3                                                       \\ 
DHHN~\cite{jiang2022dhhn}           & 61.4 & 63.4 & 56.8 & 60.5 & 59.5 & 54.6 & 60.8 & 51.1 & 55.5 & 53.3 & 57.7                                                      \\ 

CMPAE~\cite{gao2023collecting}              & {64.2}                                                       & {66.4}                                                       & {59.2}                                                       & {63.3}                                                       & 62.8                                                       & {56.6}                                                       & {63.7}                                                       & {51.8}                                                       & {57.4}                                                       & 55.7                                                       & 60.1                                                       \\ 
LSLD~\cite{fan2023revisit}                      & {62.7}                                                       & {67.1}                                                       & {59.4}                                                       & {63.1}                                                       & 62.2                                                       & {55.7}                                                       & {64.3}                                                       & {52.6}                                                       & {57.6}                                                       & 55.2                                                       & 60.0                                                       \\ 
VALOR~\cite{lai2023modality}                    & {62.8}                                                       & {66.7}                                                       & {60.0}                                                       & {63.2}                                                       & 62.3                                                       & {57.1}                                                       & {63.9}                                                       & {54.4}                                                       & {58.5}                                                       & 55.9                                                       & 60.5                                                       \\ 
\textbf{MM-CSE (ours)}                  & \textbf{65.0}                                                       & \textbf{66.8}                                                    &\textbf{60.0}                                                       & \textbf{63.9}                                                       & \textbf{64.2}                                                       & \textbf{59.1}                                                       & \textbf{64.1}                                                       & \textbf{54.9}                                                       & \textbf{59.4}                                                       & \textbf{57.6}                                                       & \textbf{61.5}                                                       \\ \hline
VALOR*~\cite{lai2023modality}                   & {68.1}                                                       & {68.4}                                                       & {61.9}                                                       & {66.2}                                                       & 66.8                                                       & {61.2}                                                       & {64.7}                                                       &{55.5}                                                       & {60.4}                                                       & 59.0                                                       & 63.2                                                       \\ 
\textbf{MM-CSE* (ours)}                  & \textbf{{\begin{tabular}[c]{@{}c@{}}69.5\\      (+1.4)\end{tabular}}} & \textbf{{\begin{tabular}[c]{@{}c@{}}71.3\\      (+2.9)\end{tabular}}} & \textbf{{\begin{tabular}[c]{@{}c@{}}64.2\\      (+2.3)\end{tabular}}} & \textbf{{\begin{tabular}[c]{@{}c@{}}68.3\\      (+2.1)\end{tabular}}} & \textbf{\begin{tabular}[c]{@{}c@{}}68.9\\      (+2.1)\end{tabular}} & \textbf{{\begin{tabular}[c]{@{}c@{}}63.0\\      (+1.8)\end{tabular}}} & \textbf{{\begin{tabular}[c]{@{}c@{}}67.5\\      (+2.8)\end{tabular}}} & \textbf{{\begin{tabular}[c]{@{}c@{}}57.8\\      (+2.3)\end{tabular}}} & \textbf{{\begin{tabular}[c]{@{}c@{}}62.7\\      (+2.3)\end{tabular}}} & \textbf{\begin{tabular}[c]{@{}c@{}}61.1\\      (+2.1)\end{tabular}} & \textbf{\begin{tabular}[c]{@{}c@{}}65.4\\      (+2.2)\end{tabular}} \\ 
\Xhline{1.5pt}
\end{tabular}
\caption{Comparison with state-of-the-art methods. The results shown in the upper Table are obtained by using audio and visual features respectively extracted by VGGish and ResNet models, while results in the last two rows ($*$) denote that the CLAP and CLIP features are used. `Avg.' is the average result of all ten metrics.}
\label{table:sota_comparison}
\end{table*}

\subsection{Comparison with State-of-the-Arts}
We quantitatively compare our proposed MM-CSE method with prior works.
Notably, previous works typically utilize VGGish~\cite{hershey2017vggish}, pretrained on AudioSet~\cite{gemmeke2017audioset} to extract audio features and employ the pretrained ResNet-152~\cite{he2016resnet} and R(2+1)D~\cite{tran2018closer} models to extract visual features.
For a fair comparison, we employ the same audio and visual backbones for feature extraction, reporting the performances in the upper part of Table~\ref{table:sota_comparison}.
Our method sets a new benchmark with a remarkable 61.5\% average parsing performance, surpassing the prior SOTA VALOR~\cite{lai2023modality}, especially in audio event parsing with a notable $\sim$2\% gain in both segment-level and event-level metrics. 
Furthermore, we compare our method against VALOR using the same CLAP and CLIP features.
As presented in the final two rows of Table~\ref{table:sota_comparison}, our method continues to outperform VALOR in all types of event parsing, surpassing it by 2.2\% on the average performance.
These results demonstrate the superiority of our proposed method. In the subsequent section, we validate the effectiveness and advantages of each core component.
\begin{table}[!t]
\small
  \centering
  \setlength{\tabcolsep}{2mm}{} 
\begin{tabular}{c|cc|ccccc|c}
\Xhline{1.5pt}
\#& EVE. &  BG  & A    & V    & AV   & Type & Event &  Avg.                     \\ \hline
1 & $\times$    & $\times$          & 64.4 & 66.2 & 58.6 & 63.1 & 62.5 & 63.0                  \\
2 & $$\checkmark$$    & $\times$          & 65.2 & 68.1 & 59.8 & 64.3 & 63.7 & 64.2                 \\
3 & $\checkmark$    & $\checkmark$         & \textbf{66.3} & \textbf{69.4} & \textbf{61.0} & \textbf{65.5} & \textbf{65.0} & \textbf{65.4} \\
\Xhline{1.5pt}
\end{tabular}
\caption{ Ablation study on the CAFD module. `EVE.' denotes $K$ event-specific classes, while `BG' stands for an additional background class.
}
\label{table:ablation_CAFD}
\end{table}

\begin{table}[!t]
\small
  \centering
\setlength{\tabcolsep}{1mm}{} 
\begin{tabular}{cc|ccccc|c}
\Xhline{1.5pt}
\multicolumn{2}{c|}{Strategy} & A    & V    & AV   & Type & Event & Avg.\\ \hline
\multicolumn{2}{c|}{full}  & \textbf{66.3} & \textbf{69.4} & \textbf{61.0} & \textbf{65.5} & \textbf{65.0} & \textbf{65.4}                              \\ \hline
\multirow{2}{*}{blocks} & w/o SECM                                      & 62.7          & 65.3          & 57.1          & 61.7          & 61.5          & 61.6                                           \\ 
 & w/o LGSF                                      & 64.5          & 66.8          & 59.0          & 63.4          & 62.6          & 63.2                                          \\
\hline
\multirow{2}{*}{modes} & w/o intra-FGSE                                & 64.6          & 68.5          & 59.6          & 64.2          & 63.5          & 64.1                                       \\
& w/o cross-FGSE                               & 59.7          & 67.2          & 54.3          & 60.4          & 61.6          & 60.6  \\
\Xhline{1.5pt}
\end{tabular}
\caption{Ablation study on the FGSE module. We explore the impacts of each block and each mode.}
\label{table:ablation_FGSE}
\end{table}

\subsection{Ablation Studies}
In this section, we conduct ablation studies to validate the key designs in our model. \textit{Notably, due to space limitations, we present the average result of segment-level and event-level metrics for the experiments.} More detailed results of each metric are further provided in our Supplementary Material.

\noindent\textbf{Ablations on the CAFD module.}
We validate this module by exploring various feature decoupling strategies, as illustrated in Table~\ref{table:ablation_CAFD}:
{\#1)} We implement our model directly using the semantically mixed features without decoupling;
{\#2)} We decouple the features into only $K$ event-specific classes (`EVE.').
{\#3)} We decouple the features into both $K$ event-specific classes and an extra background class (`EVE.'+`BG').
Our experiments reveal that the model exhibits the poorest performance 
when removing the CAFD module (\#1).
However, feature decoupling for event classes leads to a notable enhancement by an average of 1.2\% (\#2).
Additionally, optimal performance is achieved by further including the background class (\#3).
These findings suggest that it is crucial to consider feature decoupling for both the event-specific and background classes.
The decoupled features ensure more nuanced semantic modeling via class-level interactions.

\begin{table}[!t]\small
  \centering
\setlength{\tabcolsep}{1mm}{}
\begin{tabular}{c|cccc|ccccc|c}
\Xhline{1.5pt}
\# & $\mathcal{L}_\text{basic}$ &  $\mathcal{L}_\text{rec}$                    &         $\mathcal{L}_\text{ort}$               &               $\mathcal{L}_\text{ec}$        
& A             & V             & AV            & Type          & Event         &    Avg.                   \\
\hline
1 & $$\checkmark$$     & $\times$     & $\times$       & $\times$    & 65.6          & 67.8          & 59.8          & 64.4          & 64.1          & 64.3                   \\
2 & $$\checkmark$$     & $$\checkmark$$    & $\times$       & $\times$    & 65.5          & 68.1          & 59.4          & 64.3          & 64.1          & 64.3                  \\
3 & $$\checkmark$$     & $\times$    &   $$\checkmark$$    & $\times$    & 65.3          & 68.4          & 59.6          & 64.4          & 64.3          & 64.4
\\
4 & $$\checkmark$$     & $$\checkmark$$    & $$\checkmark$$      & $\times$    & 65.8          & 69.0          & 60.0          & 64.9          & 64.7          & 64.8                   \\
5 & $\checkmark$    & $\checkmark$     & $\checkmark$       & $\checkmark$    & \textbf{66.3} & \textbf{69.4} & \textbf{61.0} & \textbf{65.5} & \textbf{65.0} & \textbf{65.4}    \\ \Xhline{1.5pt}
\end{tabular}
\caption{Ablation study on the loss functions.}
\label{table:ablation_losses}
\end{table}

\begin{figure}[!t]
  \centering
  \includegraphics[width=1\linewidth]{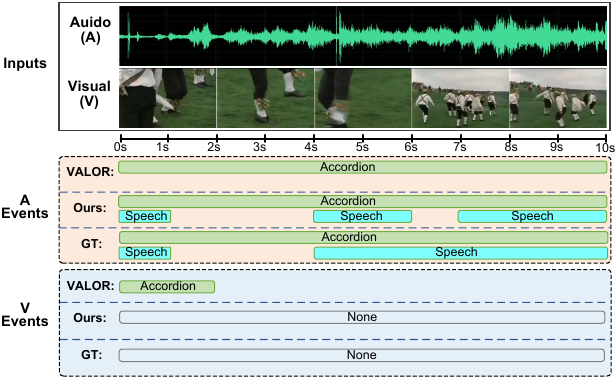}
  \caption{
  Qualitative comparisons. Compared to the previous SOTA method VALOR, our method performs better in identifying multiple overlapping events and recognizing or utilizing the background information.
  }
  \label{fig:avvp_examples}
\end{figure}

\noindent\textbf{Ablations on the FGSE module.}
As introduced in Sec.~\ref{sec:FGSE}, the proposed FGSE module comprises two blocks: the segment-wise co-occurrence modeling (SECM) and the local-global semantic fusion (LGSF).
Besides, the FGSE operates in two modes: intra-modality and cross-modality.
We perform ablations on each block and each mode to assess their impacts.
As shown in Table~\ref{table:ablation_FGSE}, the model's performance experiences a notable decrease when either block or mode is removed.
This demonstrates the effectiveness of each component of the FGSE module.
From the last two rows of the Table, we can also conclude that considering the cross-modal interactions is more important for the AVVP task.

\noindent\textbf{Ablations on the loss functions.}
We explore the impact of each loss by ablating them to train the model. 
As shown in Table~\ref{table:ablation_losses}, the model trained solely with the basic $\mathcal{L}_\text{basic}$ already achieves satisfactory performance, with an average metric of 64.3\% (\#1).
Notably, our model trained in this setting already surpasses the previous SOTA method VALOR (63.2\%), demonstrating the superiority of our network design.
Moreover, it is observed that using $\mathcal{L}_\text{rec}$ and $\mathcal{L}_\text{ort}$ independently does not yield obvious benefits (\#2 and \#3).
However, combining these two losses enhances the performance, suggesting their complementarity for effective class-aware feature decoupling (\#4).
Finally, the introduction of $\mathcal{L}_\text{ec}$ further improves the performance by 0.6\% as it provides additional supervision during event co-occurrence modeling (\#5).

\begin{figure}[!t]
  \centering
  \includegraphics[width=0.95\linewidth]{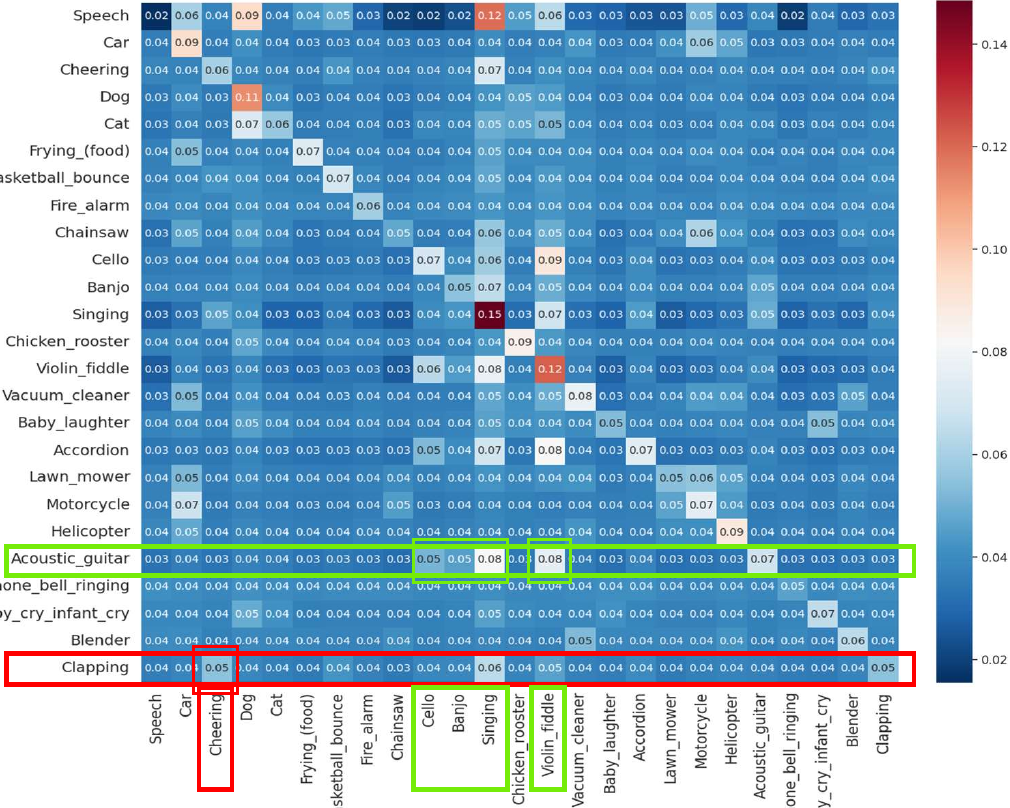}
  \caption{Visualization example of the learned event co-occurrence map. }
  \label{fig:event_cooccurrence_map}
\end{figure}

\subsection{Qualitative Results}
\textbf{Qualitative example of audio-visual video parsing.}
We first present a parsing example to intuitively compare our proposed method with the previous SOTA method VALOR.
As shown in Fig.~\ref{fig:avvp_examples}, there are two audio events: \textit{Accordion} and \textit{Speech}.
VALOR successfully identifies \textit{Accordion} event but completely misses the \textit{Speech} event.
Conversely, our method satisfactorily localizes segments containing both events.
An interesting observation is evident in the visual track where VALOR incorrectly recognizes that the event \textit{Accordion} occurs in the first two segments.
However, the visual frames actually depict people's feet on the playground.
This misunderstanding may arise from semantic interference from the audio events related to the \textit{Accordion}, as VALOR performs cross-modal interaction directly relying on semantically mixed audio and visual features.
In contrast, our method accurately classifies the first two segments as backgrounds.
Our method utilizes decoupled class-aware features for intra- and cross-modal interactions, enabling each class's feature to aggregate only similar semantics from relevant classes.
Additionally, introducing an additional background class during the feature decoupling process helps the model distinguish the background from diverse events, preventing the model from consistently falling into predefined event categories. 

\noindent\textbf{Visualization of learned event co-occurrence map.} We then present the learned event co-occurrence map, derived by averaging the $\bm{\beta}^{av} \in \mathbb{R}^{T \times K \times K}$ across all test videos along the temporal dimension.
In Fig.~\ref{fig:event_cooccurrence_map}, the visualization illustrates that our model effectively captures the co-occurrence among relevant events.
For instance, as highlighted by the green boxes, the audio event \textit{Acoustic\_guitar} exhibits higher correspondence with the visual event \textit{Singing} and some instrumental events.
Similarly, the audio event \textit{Clapping} correlates well with the visual event 
\textit{Cheering} (red boxes). 
These observations align with common scenarios where such audio-visual event pairs frequently co-occur.
Modeling such event co-occurrence enhances the model's capability to perceive concurrent events.

\begin{figure}[!t]
  \centering
  \includegraphics[width=1\linewidth]{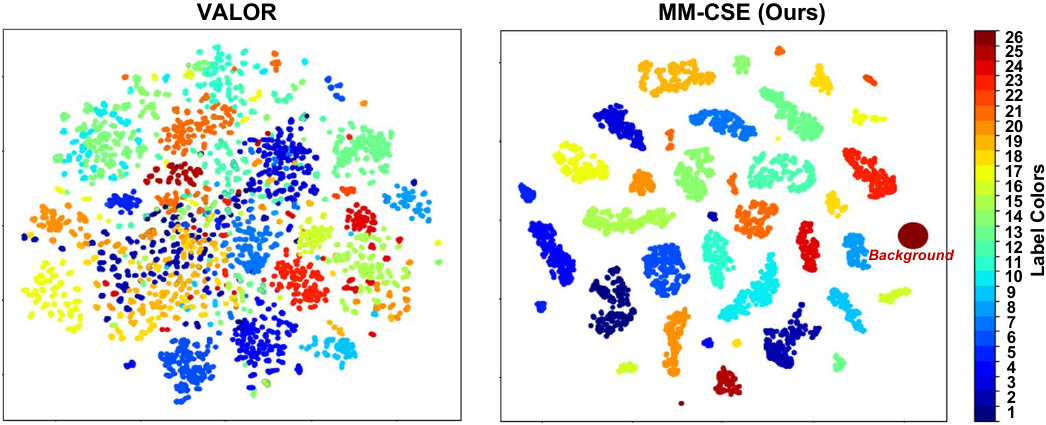}
  \caption{Visualization of our decoupled class-wise features. 
  Each color represents one class.}
  \label{fig:tsne}
\end{figure}

\noindent\textbf{Visualization of the decoupled class-wise features.}
Finally, we visualize the distributions of our decoupled features using the t-SNE~\cite{van2008visualizing}.
Here, we take the audio modality for demonstration.
In Fig.~\ref{fig:tsne}, we also compare our method with the previous SOTA method VALOR which exhibits semantically mixed audio representations across different event classes. In contrast, our decoupled features, including both event-specific and background features, demonstrate pronounced intra-class compactness and inter-class separation. These results validate that our method effectively decouples the semantically mixed hidden features into separate class-aware features. 
Such class-wise features allow more accurate interactions from class-level for both intra- and cross-modality, thereby enhancing multi-class event classification in the AVVP task.

\section{Conclusion}
For the audio-visual video parsing task, existing
approaches directly utilize semantically mixed holistic audio and visual features for modeling intra- and cross-modal relations, which could result in semantic interference.
In this paper, we propose decoupling the semantically mixed features into distinct event-specific and background-specific class-wise features.
These decoupled features allow for precise event semantic learning for each segment by aggregating positive supports from features of highly relevant classes. 
Specifically, the proposed FGSE 
module first encodes the inter-class dependencies among concurrent events within each timestamp, and then enriches each local segment with matched global video semantics.
Additionally, several loss functions are introduced for effective feature decoupling and event co-occurrence modeling.
Extensive experiments verify the superiority of our class-aware semantic enhancement method.

\section*{Acknowledgements}
We would like to express our sincere gratitude to the anonymous reviewers for their invaluable comments and insightful suggestions.
This work was supported by the National Natural Science Foundation of China (61972127, 62272142, 62272144, 72188101, and 62020106007), the Major Project of Anhui Province (2408085J040, 202203a05020011), and the Fundamental Research Funds for the Central Universities (JZ2024HGTG0309, JZ2024AHST0337, and JZ2023YQTD0072).

\bibliography{aaai25}


\newpage
\section{Supplementary Material}

In this section, we initially discuss the computational efficiency of the proposed MM-CSE network in Sec.~\ref{Supp_Computational Efficiency}. Subsequently, in Sec.~\ref{Supp_abla_compar}, we give some additional experimental results. Finally, we present some audio-visual video parsing examples in Sec.~\ref{Supp_examples}. Besides, detailed results of each metric across all ablation studies in the main paper are listed in Tables~\ref{table:ablation_CAFD_supp},~\ref{table:ablation_FGSE_supp}, and~\ref{table:ablation_losses_supp}.

\subsection{Analysis of Computational Efficiency}\label{Supp_Computational Efficiency}
As depicted in Table~\ref{table:Computational efficiency_supp}, we compare our method with the prior state-of-the-art method VALOR~\cite{lai2023modality} in terms of model efficiency. It can be found that: 
1) Our model possesses \textit{fewer} parameters compared to VALOR, despite the additional parameters introduced by the CAFD module. This difference arises because the hidden dimension of our FGSE module is 128, whereas it is always 256 in the VALOR network. 
2) Our model has higher FLOPs and Runtime 
because our FGSE module operates on fine-grained class-wise features; nevertheless, it remains \textit{affordable}. 
3) Importantly, despite this trade-off, our model achieves a \textit{significant average performance improvement} (2.2\%) compared to VALOR.

\begin{table}[h]

  \centering
\setlength{\tabcolsep}{1.5mm}{} 
\begin{tabular}{c|ccc}
\Xhline{1.5pt}
Methods & \#Params.(M)$\downarrow$  & FLOPs(G)$\downarrow$   & Runtime(ms)$\downarrow$   \\ \midrule
VALOR       & 5.05 & 0.45 & 7.34  \\
\textbf{Ours}      & \textbf{4.49} & 0.80 & 12.63 
\\ 
\Xhline{1.5pt}
\end{tabular}
\vspace{-1ex}
\caption{Analysis of computational efficiency.}
\label{table:Computational efficiency_supp}
\vspace{-3ex}
\end{table}

\subsection{Additional Ablation Study and Comparisons}\label{Supp_abla_compar}
\textbf{Ablation study on the weights of different $\lambda_1$ and $\lambda_2$ in Eq.~\ref{eq:loss_total}.} 
As can be seen from Table~\ref{table:ablation_weights_supp}, the model performance can be influenced by these parameters for balancing the loss items. 
Based on the best-case weight here i.e. 0.1, we carried out the experiments in this paper.

\begin{table*}[h]
  \centering
\setlength{\tabcolsep}{2mm}
\begin{tabular}{c|cc|ccccc|ccccc|c}
\Xhline{1.5pt}
\multirow{2}{*}{\#}& \multirow{2}{*}{$\lambda_1$} & \multirow{2}{*}{$\lambda_2$} & \multicolumn{5}{c|}{Segment-level}                                             & \multicolumn{5}{c|}{Event-level}                                             & \multirow{2}{*}{Avg.} \\
                    &     &                          & A             & V             & AV            & Type@AV          & Event@AV          & A           & V             & AV            & Type@AV           & Event@AV          &                      \\
                         \midrule
1 & \textbf{0.1}             & \textbf{0.1}             & \textbf{69.5} & \textbf{71.3} & \textbf{64.2} & \textbf{68.3} & \textbf{68.9} & \textbf{63} & \textbf{67.5} & \textbf{57.8} & \textbf{62.7} & \textbf{61.1} & \textbf{65.4}        \\
2 &0.5                      & 0.5                      & 68.4          & 70.0          & 63.1          & 67.2          & 67.7          & 61.0        & 66.5          & 56.7          & 61.4          & 59.6          & 64.2                 \\
3 & 1.0                      & 1.0                      & 69.5          & 70.1          & 63.3          & 67.6          & 68.7          & 62.3        & 65.9          & 56.4          & 61.5          & 59.9          & 64.5     \\ \Xhline{1.5pt}       
\end{tabular}
  \caption{Ablation study on the weights of different $\lambda_1$ and $\lambda_2$ in Eq.~\ref{eq:loss_total}.}
  \label{table:ablation_weights_supp}
\end{table*}

\begin{table*}[h]
  \centering
\setlength{\tabcolsep}{1.2mm}
\begin{tabular}{c|ccccc|ccccc|c}
\Xhline{1.5pt} 
\multirow{2}{*}{Methods} & \multicolumn{5}{c|}{Segment-level}                                    & \multicolumn{5}{c|}{Event-level}                                      & \multirow{2}{*}{Avg.} \\
                         & A             & V             & AV   & Type@AV          & Event@AV          & A             & V             & AV   & Type@AV           & Event@AV          &                      \\ \midrule
LFAV~\cite{hou2024toward}                     & 60.5          & 64.6          & 57.5          & 60.9          & 59.6          & 52.3          & 61.9          & 50.6          & 54.9          & 51.5          & 57.4                 \\
\textbf{MM-CSE(ours)}            & \textbf{65.0} & \textbf{66.8} & \textbf{60.0} & \textbf{63.9} & \textbf{64.2} & \textbf{59.1} & \textbf{64.1} & \textbf{54.9} & \textbf{59.4} & \textbf{57.6} & \textbf{61.5} \\ \midrule
LEAP~\cite{zhou2025label}$\dag$                   & 64.8          & 67.7          & 61.8 & 64.8          & 63.6          & 59.2          & 64.9          & 56.5 & 60.2          & 57.4          & 62.1                 \\
\textbf{MM-CSE(ours)$\dag$}            & \textbf{65.2} & \textbf{68.3} & 61.4 & \textbf{65.0} & \textbf{64.6} & \textbf{59.5} & \textbf{65.4} & 56.4 & \textbf{60.4} & \textbf{58.0} & \textbf{62.4}       \\
\Xhline{1.5pt}  
\end{tabular}
  \caption{Comparisons with some related works - LFAV and LEAP, where `$\dag$' denotes the models use the same front-end backbone.}
  \label{table:compar_LEAP_LFAV_supp}
\end{table*}

\begin{figure*}[!t]
  \centering
  \includegraphics[width=0.95\linewidth]{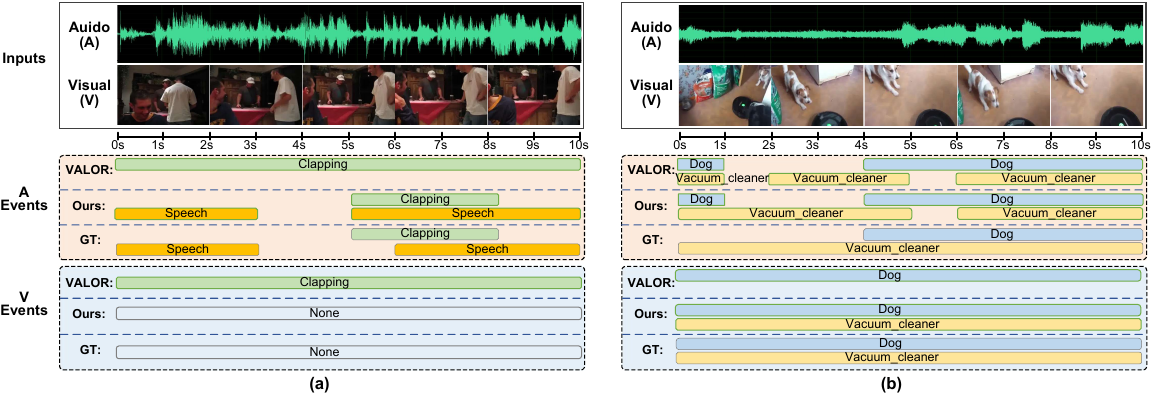}
  \caption{Qualitative examples of video parsing. Compared to the previous SOTA method VALOR, our method performs better in identifying multiple overlapping events and recognizing or utilizing the background information. 
  }
  \label{fig:avvp_examples_supp}
\end{figure*}

\begin{table*}[!t]
  \centering
  \setlength{\tabcolsep}{2mm}{} 
\begin{tabular}{c|cc|ccccc|ccccc|c}
\Xhline{1.5pt}
\multirow{2}{*}{\#} & \multicolumn{2}{c|}{Class type}    & \multicolumn{5}{c|}{Segment-level}       & \multicolumn{5}{c|}{Event-level}         & \multirow{2}{*}{Avg.} \\
& EVE. &  BG  & A    & V    & AV   & Type@AV & Event@AV & A    & V    & AV   & Type@AV & Event@AV &                       \\ \midrule
1 & $\times$    & $\times$          & 67.8 & 68.3 & 61.9 & 66.0    & 66.6     & 61.0 & 64.1 & 55.3 & 60.2    & 58.4     & 63.0                  \\
2 & $$\checkmark$$    & $\times$          & 68.5 & 70.0 & 62.9 & 67.1    & 67.6     & 61.8 & 66.2 & 56.6 & 61.5    & 59.7     & 64.2                  \\
3 & $\checkmark$    & $\checkmark$         & \textbf{69.5} & \textbf{71.3} & \textbf{64.2} & \textbf{68.3}    & \textbf{68.9}     & \textbf{63.0} & \textbf{67.5} & \textbf{57.8} & \textbf{62.7}    & \textbf{61.1}     & \textbf{65.4} \\
\Xhline{1.5pt}

\end{tabular}
\vspace{-1ex}
\caption{ Ablation study on the CAFD module.
`EVE' indicates that the semantically mixed audio/visual features are decoupled into $K$ event-specific classes, while `BG' means that we further consider an additional background class.}
\label{table:ablation_CAFD_supp}
\end{table*}

\begin{table*}[!t]
  \centering
  \setlength{\tabcolsep}{1.5mm}{} 
\begin{tabular}{cc|ccccc|ccccc|c}
\Xhline{1.5pt}
\multicolumn{2}{c|}{\multirow{2}{*}{Strategy}} & \multicolumn{5}{c|}{Segment-level} & \multicolumn{5}{c|}{Event-level} & \multicolumn{1}{c}{\multirow{2}{*}{Avg.}} \\
& &A    & V    & AV   & Type@AV & Event@AV & A    & V    & AV   & Type@AV & Event@AV & \\ \midrule
\multicolumn{2}{c|}{full}  & \textbf{69.5} & \textbf{71.3}   & \textbf{64.2} & \textbf{68.3}             & \textbf{68.9}              & \textbf{63.0}          & \textbf{67.5}          & \textbf{57.8}           & \textbf{62.7}             & \textbf{61.1}              & \textbf{65.4}                              \\ \midrule
\multirow{2}{*}{blocks} & w/o SECM                                      & 68.1                   & 68.9                   & 62.5                    & 66.5                      & 67.0                       & 60.8                   & 64.6                   & 55.5                    & 60.3                      & 58.1                       & 63.2                                       \\ 
 & w/o LGSF                                      & 66.8                   & 68.5                   & 61.1                    & 65.5                      & 66.5                       & 58.5                   & 62.1                   & 53.0                    & 57.9                      & 56.4                       & 61.6                                       \\
\midrule
\multirow{2}{*}{modes} & w/o intra-FGSE                                & 68.0                  & 70.5                  & 62.9                   & 67.1                     & 67.5                      & 61.2                  & 66.6                  & 56.2                   & 61.3                     & 59.4                      & 64.1                                      \\
& w/o cross-FGSE                               & 63.5                  & 69.1                  & 57.3                   & 63.3                     & 65.8                      & 55.8                  & 65.3                  & 51.2                   & 57.5                     & 57.3                      & 60.6 \\
\Xhline{1.5pt}
\end{tabular}
\caption{Ablation study on the FGSE module. We explore the impacts of each block and each mode.}
\label{table:ablation_FGSE_supp}
\end{table*}

\begin{table*}[!t]
  \centering
\setlength{\tabcolsep}{1.2mm}
\begin{tabular}{c|cccc|ccccc|ccccc|c}
\Xhline{1.5pt}
\multirow{2}{*}{\#}& \multirow{2}{*}{$\mathcal{L}_\text{basic}$} & \multirow{2}{*}{$\mathcal{L}_\text{rec}$} & \multirow{2}{*}{$\mathcal{L}_\text{ort}$} & \multirow{2}{*}{$\mathcal{L}_\text{ec}$}                                                            & \multicolumn{5}{c|}{Segment-level}                                                                                    & \multicolumn{5}{c|}{Event-level}                                                                                      & \multirow{2}{*}{Avg.} \\ 
 & &                      &                        &                       
& A             & V             & AV            & Type@AV          & Event@AV         & A             & V             & AV            & Type@AV          & Event@AV         &                       \\
\midrule
1 & $$\checkmark$$     & $\times$     & $\times$       & $\times$    & 69.0          & 69.5          & 62.9          & 67.1          & 67.7          & 62.2          & 66.0          & 56.6          & 61.6          & 60.4          & 64.3                  \\
2 & $$\checkmark$$     & $$\checkmark$$    & $\times$       & $\times$    & 68.6          & 70.1          & 62.7          & 67.1          & 67.8          & 62.3          & 66.1          & 56.1          & 61.5          & 60.3          & 64.3                  \\
3 & $$\checkmark$$     & $\times$    &   $$\checkmark$$    & $\times$    & 68.9          & 70.1          & 62.7          & 67.2          & 68.3          & 61.6          & 66.6          & 56.4          & 61.6          & 60.2          & 64.4
\\
4 & $$\checkmark$$     & $$\checkmark$$    & $$\checkmark$$      & $\times$    & 69.4          & 70.8          & 63.4          & 67.9          & 68.7          & 62.1          & 67.1          & 56.5          & 61.9          & 60.6          & 64.8                  \\
5 & $\checkmark$    & $\checkmark$     & $\checkmark$       & $\checkmark$    & \textbf{69.5} & \textbf{71.3} & \textbf{64.2} & \textbf{68.3} & \textbf{68.9} & \textbf{63.0} & \textbf{67.5} & \textbf{57.8} & \textbf{62.7} & \textbf{61.1} & \textbf{65.4}    \\ \Xhline{1.5pt}
\end{tabular}
  \caption{Ablation study on the loss functions.}
  \label{table:ablation_losses_supp}
\end{table*}

\noindent\textbf{Comparisons with More Parsing Works - LFAV~\cite{hou2024toward} and LEAP~\cite{zhou2025label}.}
1) LFAV focuses on the task of long-form video parsing, and still rely on semantically mixed holistic features without considering the issue of semantic interference. We evaluate LFAV on the LLP dataset. As shown in the upper part of Table~\ref{table:compar_LEAP_LFAV_supp}, our method significantly
outperforms it. 
2) LEAP, the latest work, achieves event disentanglement by projecting holistic features onto additional preprocessed class text features, while we directly generate class-aware, decoupled features from the holistic feature. Note that the optimal LEAP model uses the MM-Pyr~\cite{yu2021mm} as the backbone. As shown in the lower part of Table~\ref{table:compar_LEAP_LFAV_supp}, if adopting the same setup, our model is still competitive with LEAP.

\noindent\textbf{Discussion with More Related Works - UniAV~\cite{geng2024uniav} and ETAVR~\cite{senocak2023event}.} 
There are differences in tasks.
UniAV focuses on designing a unified framework to address three tasks—sound event detection, temporal action localization, and audio-visual event localization.
And ETAVR attempts to improve video-level classification by subdividing five event types (i.e., audio-only, visual-only, continuous, onset, and instant events), whereas we address the segment-level AVVP task by using
multiple linear layers for decoupling mixed event semantics. Additionally, ETAVR achieves the F1 score of 72\% on the LLP dataset, compared to 88\% for our model.

\subsection{More Examples of Audio-Visual Video Parsing}\label{Supp_examples}
To illustrate our method's superiority more intuitively, we present some audio-visual video parsing examples. As shown in Fig.~\ref{fig:avvp_examples_supp} (a), our method satisfactory identifies the \textit{`Speech'} event in addition to the audio event \textit{`Clapping'}, where VALOR fails. Furthermore, in the visual modality, VALOR incorrectly recognizes the background as the \textit{`Clapping'} event. This error arises because VALOR performs interactions based on semantically mixed holistic audio and visual features, leading to semantic interference. In contrast, our method effectively mitigates this issue through the proposed class-aware semantic enhancement.
Likewise, in Fig.~\ref{fig:avvp_examples_supp} (b), our method provides more accurate predictions for both audio and visual events. Notably, while VALOR only identifies the visual event \textit{`Dog'}, our model successfully localizes the \textit{`Vacuum cleaner'}, despite it being partially obscured in several segments. This improvement may be attributed to our method's consideration of decoupled background information (\textit{`floor'}), which is related to the event \textit{`Vacuum cleaner'}, thereby enhancing event recognition.

\end{document}